\documentclass[conference]{IEEEtran}
\IEEEoverridecommandlockouts

\usepackage{cite}
\usepackage{amsmath,amssymb,amsfonts}
\usepackage{algorithm}
\usepackage{algorithmic}
\usepackage{graphicx}
\usepackage{caption}
\usepackage{textcomp}
\usepackage{xcolor}
\usepackage{booktabs} 
\usepackage{multirow} 
\usepackage{cuted}
\usepackage[figuresright]{rotating}

\def\BibTeX{{\rm B\kern-.05em{\sc i\kern-.025em b}\kern-.08em
    T\kern-.1667em\lower.7ex\hbox{E}\kern-.125emX}}
\begin{document}

\title{$\text{M}^2$DQN: A Robust Method for Accelerating Deep Q-learning Network
}

\author{
\IEEEauthorblockN{
  Zhe Zhang\textsuperscript{1}\IEEEauthorrefmark{1},
  Yukun Zou\textsuperscript{2}\IEEEauthorrefmark{2}\IEEEauthorrefmark{3},
  Junjie Lai\textsuperscript{3}\IEEEauthorrefmark{2}\IEEEauthorrefmark{3},
  Qing Xu\textsuperscript{4}\IEEEauthorrefmark{1}
 }
 \IEEEauthorblockA{\IEEEauthorrefmark{1}UniDT, Shanghai, China}
 \IEEEauthorblockA{\IEEEauthorrefmark{2}Research Institute of Intelligent Complex Systems, Fudan University, Shanghai, China}
 \IEEEauthorblockA{\IEEEauthorrefmark{3}Institute of Science and Technology for Brain-Inspired Intelligence, Fudan University, Shanghai, China}
 \IEEEauthorblockA{\textsuperscript{1}zhe.zhang@unidt.com, \textsuperscript{2}ykzou19@fudan.edu.cn, \textsuperscript{3}jjlai19@fudan.edu.cn, \textsuperscript{4}qing.xu@unidt.com}
 \IEEEauthorblockA{Zhe Zhang, Yukun Zou, Junjie Lai contributed equally to this work}
 }

\maketitle

\begin{abstract}
Deep Q-learning Network (DQN) is a successful way which combines reinforcement learning with deep neural networks and leads to a widespread application of reinforcement learning. One challenging problem when applying DQN or other reinforcement learning algorithms to real world problem is data collection.
Therefore, how to improve data efficiency is one of the most important problems in the research of reinforcement learning.
In this paper, we propose a framework which uses the Max-Mean loss in Deep Q-Network ($\text{M}^2$DQN).
Instead of sampling one batch of experiences in the training step, 
we sample several batches from the experience replay and update the parameters such that the maximum TD-error of these batches is minimized.
The proposed method can be combined with most of existing techniques of DQN algorithm by replacing the loss function.
We verify the effectiveness of this framework with one of the most widely used techniques, Double DQN (DDQN), in several gym games.
The results show that our method leads to a substantial improvement in both the learning speed and performance.
\end{abstract}

\begin{IEEEkeywords}
deep Q network, data efficiency, mini-max, robust optimization
\end{IEEEkeywords}

\section{Introduction}
In reinforcement learning (RL), the agent explores and learns by interacting with the environment, with the ultimate goal of maximizing cumulative reward \cite{rl}. In recent years, a variety of different reinforcement learning algorithms have been applied to classic challenging tasks, such as Go \cite{go}, Atari Games \cite{dqn}, robotics \cite{robot}, etc., which have achieved excellent performance. In this process, Deep Q-Network (DQN) \cite{dqn} and its variants, such as Double DQN \cite{doubleDQN}, Dueling DQN \cite{duelDQN}, play an important role.

One of the most important techniques in DQN is the use of experience replay \cite{experience_replay}.
The experience replay address the issues of temporal correlation between transitions and data reuse.
Various studies and improvements of experience replay have been proposed.
To reduce the forgetting in long-time training, Zhang et al. (2019) \cite{dual_experience_replay} introduce a framework of dual replay buffer.
To improve the data efficiency, Schaul et al. (2016) \cite{PER} introduce prioritized experience replay (PER), in which the transitions are prioritized by the last encountered absolute TD error. Cao et al. (2019) \cite{HVPER} design a kind of PER based on both reward and TD error, leading to  further improvement on training efficiency.
Another widely used technique in DQN is the use of target network, which stabilizes the learning of DQN by reducing variance.
Experience replay and target network have become standard practice in DQN algorithms to achieve state-of-the-art performance.

In this paper, we propose a new method to improve the data efficiency.
Instead of sampling one batch of experiences in the training step, we sample several batches from the experience replay and update the parameters such that the maximum TD-error of these batches is minimized. Such maximum TD-error is called the mini-max loss, for short.

The mini-max loss is usually used in robust optimization, generative adversarial learning and worst-case optimization. The mini-max loss treat exceptional samples and the normal samples in an equal way, and fully consider the influence of all examples, especially the worst-case samples. Thus, the mini-max loss helps to learn a robust model. 

We develop a framework which combines the Mini-Max method with Deep
Q-Network and its variants (We call it $\text{M}^2$DQN). We evaluate our algorithm with Double DQN (DDQN) on several gym \cite{gym} environments in this paper.
The results show that the proposed method leads to a substantial improvement in both the learning speed and performance.

\section{Background}
Reinforcement learning addresses the task of an agent learning to interact with the environment in order to maximize the cumulative future reward. 
We model this interaction as a discounted Markov Decision Process (MDP) $(\mathcal{S}, \mathcal{A}, T, R, \gamma)$, which consists of states $\mathcal{S}$, actions $\mathcal{A}$, a reward function $R(s,a): S\times A \rightarrow \mathbb{R}$, a state transition function $T(s,a,s') = P(s' \mid s, a)$ and a discounted factor $\gamma \in [0,1]$.
A policy $\pi$ is a function that maps every state $s \in \mathcal{S}$ to a probability distribution over the action space.

The discounted cumulative future reward (discounted return) $R_t$ at time-step $t$ is usually defined to be the discounted sum of future rewards 

$$R_t = \sum_{k=0}^{\infty} \gamma^k r_{t+k}$$

\noindent where $r_t$ is the reward at time-step $t$.
Under a given policy $\pi$, the action-value function is defined as the expected return starting from state $s$ and action $a$, i.e.,

$$
Q^{\pi}(s, a) = \mathbb{E}_{\pi}\left[R_t \mid S_t=s, A_t=a\right]
$$

\noindent we define the optimal action-value function $Q^*(s, a)$ as follows

$$
Q^*(s, a) = \max_{\pi} Q^{\pi}(s, a)
$$

The goal of reinforcement learning is to find an optimal policy that maximizes the expected discounted return.
Note that an optimal policy can be derived from the optimal action-values by selecting the the action of maximum value in each state.
Therefore, the goal becomes to find an optimal action-value function $Q^*(s, a)$.
The optimal action-values $Q^*(s,a)$ obeys the Bellman equation \cite{bellman}, i.e.

\[
Q^*(s, a) = \mathbb{E} \left[R(s,a) + \gamma \sum_{s'} P(s' \mid s, a) \max_{a'} Q^*(s', a')\right]
\]

\noindent and can be learned by Q-learning \cite{qlearning}.
When state and action spaces are large, it is intractable to learn action-value for each state and action pair independently.
To overcome the difficulty of large state and action spaces, deep neural network $Q(s,a,\theta)$ with weights $\theta$ is used to estimate the action-value function, which is called deep Q-network.
Then, At each time-step $t$, the agent observes a state $s_t$ of the environment and receives a reward $r_t$ following the selected action $a_t$ and resulting a state $s_{t+1}$, the parameters $\theta_t$ of Q-network $Q(s,a,\theta_t)$ can be updated as follows:

$$
\theta_{t+1} \leftarrow \theta_t + \gamma \cdot \text{TD}(s_t, a_t, r_t, s_{t+1}; \theta_t) \nabla_{\theta_t} Q(s_t, a_t, \theta_t)
$$

\noindent where

\[
\text{TD}(s_t, a_t, r_t, s_{t+1}) = r_t + \gamma \max_{a} Q(s_{t+1}, a; \theta_t) - Q(s_t, a_t; \theta_t),
\] 
\noindent which is called TD-error.

However, the learning of DQN is unstable. One of the reasons is that consecutive samples generated in the learning of DQN are highly correlated which breaks the i.i.d. principle (dependently and
identically distributed) of machine learning. Two techniques have been used to stabilize the learning of DQN.
The first one is experience replay, which holds last thousands transitions.
At training time, a batch of these transitions is sampled uniformly and used to update the network.
The use of experience replay breaks the temporal correlation between transitions.
The second one is the double Q-network, in which a separate target network is used to estimate the action-values (The target network is copied every few steps from the regular network) and the regular network is used to calculate the argmax over next states.
Separating the choosing of actions and the estimation of action-values reduces the overestimation in regular DQN.





The use of the experience replay can alleviate the non-iid problem of consecutive samples to a certain extent . In this paper, we propose a new method to solve the non-iid problem in the learning of DQN based on the nonlinear expectation theory.

The nonlinear expectation theory is laid down by Peng\cite{nonlinear}. Nonlinear expectation, including sublinear expectation as its special case, is a functional $ \hat{\mathbb{E}} : \mathcal{H} \mapsto \mathbb{R} $ satisfying monotonicity, constant preserving, sub-additivity, and positive homogeneity. It is a new framework of probability theory to characterize the uncertainty and has potential applications in some scientific fields, especially in risk management. 

In statistical learning, we define maximum expectation as same concept of the sublinear expectation $\hat {\mathbb{E}}$ ,which can be represented as the upper expectation of a subset of linear expectations , i.e.,
\begin{align*}
\hat{\mathbb{E}}(X)=\sup_{j \in J}\mathbb{E}_j(X),
\end{align*}
indexed by $ j \in J $. 

The usual estimation of expected error is based on the Law of Large Numbers. When the i.i.d. condition breaks, LLN is not applicable. However, the nonlinear LLN still holds, i.e., the limit distribution of
$$ \frac{X_1+X_2+\cdots+X_n}{n} $$
is a maximal distribution (see \cite{nonlinear}). The parameter of the maximal distribution can be estimated using a max-mean statistics (see \cite{maxmean}). 

Motivated by this work,
Xu et al. (2019) \cite{minimax} consider a class of nonlinear regression problems without the i.i.d. assumption. They split the training set into $N$ groups such that in each group the i.i.d. assumption can be satisfied. Then, the following max-mean loss is used.

$$
\max_{1 \le j \le N} \frac{1}{n_j} \sum_{k=1}^{n_j} (g^{\theta}(x_{jk}) - y_{jk})^2,
$$

\noindent where $n_j$ is the number of samples in group $j$, $(x_{jk},y_{jk})$ is the $k$-th sample in group $j$ and $g^{\theta}$ is the model function parameterized by $\theta$ which we want to learn.

We borrow this idea and propose the framework of $\text{M}^2\text{DQN}$.

\section{Method}


In this section, we formulate the Max-Mean Deep Q-learning Network ($\text{M}^2$DQN).
Instead of sampling only one batch at each training step, 
we sample $N$ batches from the experience replay at each time-step $t$.
Then the loss function of max-mean TD-error can be defined as follows

\[
\begin{aligned}
&\mathcal{L}(N;\theta) =  \max_{1 \le j \le N} \frac{1}{n_j} \sum_{k=1}^{n_j} \text{TD}(s_{tjk}, a_{tjk}, r_{tjk}, s_{t+1,jk}; \theta)^2 \\
\end{aligned}
\]

\noindent where $n_j$ is the sample size and $(s_{t, j}, a_{t,j}, s_{t+1,j}, r_{t,j})$ is the transition of $j$th batch.

Since the experience replay breaks the temporal correlations of transitions, the i.i.d condition is satisfied (or considered satisfied) in each group (batch).
Therefore, to minimize the max-mean loss, we follow the algorithm of Xu et al. (2019)\cite{minimax}.
Denote

\[
f_j(\theta) = \frac{1}{n_j} \sum_{k=1}^{n_j} \text{TD}(s_{tjk}, a_{tjk}, r_{tjk}, s_{t+1,jk}; \theta)^2
\]

\noindent and 

\[
\Phi(\theta) = \max_{1 \le j \le N} f_j(\theta)
\]

To find the descent direction at each time-step $t$, we linearize $f_j$ and $\theta_t$ and obtain the convex approximation of $\Phi$ as

\[
\hat{\Phi}(\theta) = \max_{1 \le j \le N} \{ f_j(\theta_t) + \langle \nabla f_j (\theta_t), \theta - \theta_t \rangle \}
\]

By adding a regularization term and setting $d = \theta - \theta_t$, the minimization problem becomes

\begin{equation}
\min_{d}  \max_{1 \le j \le N} \left\{ f_j(\theta_t) + \langle \nabla f_j (\theta_t), \theta - \theta_t \rangle + \frac{1}{2} \Vert d \Vert^2 \right\},
\label{dd_qp}
\end{equation}

\noindent which is equivalent to

\begin{equation}
    \min_{d,a} \frac{1}{2} \Vert d \Vert^2 + a
    \label{qp_prime1}
\end{equation}

\begin{equation}
    \text{s.t. } f_j(\theta_t) + \langle \nabla f_j (\theta_t), d \rangle \le a, \forall 1 \le j \le N. 
    \label{qp_prime2}
\end{equation}

\noindent When the dimension of $d$, i.e., the number of parameters of Q-network, is large, solving Problem (\ref{qp_prime1})-(\ref{qp_prime2}) is time-consuming.
Hence we turn to the dual problem

\begin{align}
    \min_\lambda \left(\frac{1}{2} \lambda^TGG^T\lambda - f^T\lambda\right) \label{qp_dual1}\\
    \text{s.t.} \sum_{i=1}^N \lambda_i=1, ~\lambda_i \ge 0.
    \label{qp_dual2}
\end{align}

\noindent where $G=\nabla f, f=(f_1(\theta_t), \ldots, f_N(\theta_t))^T$ is the Jacobian matrix.
Note that the dimension of the dual problem (\ref{qp_dual1})-(\ref{qp_dual2}) is $N$ (number of groups), 
which is independent of $n$ (number of parameters). 
Let $\lambda$ be a solution of the dual problem (\ref{qp_dual1})-(\ref{qp_dual2}). 
Then, $d_k = -G^T \lambda$ is the solution of problem (\ref{qp_prime1})-(\ref{qp_prime2}), which is also a descent direction.
Therefore, we update $\theta_t$ as follows

$$
\theta_{t+1} \leftarrow \theta_t - \alpha G^T\lambda 
$$

\noindent where $\alpha$ is the learning rate. 

The max-mean loss function can be combined with most of existing techniques of DQN algorithm by simply replacing the loss function. Algorithm 1 shows an example of the case of DDQN.

\begin{algorithm*}[t]
\caption{Double DQN with max-mean loss ($\text{M}^2$DDQN)}
\label{alg:Framwork}
\begin{algorithmic}[1]
\REQUIRE batchsize $K$; max step $T$; group size $N$; discount rate $\gamma$;

\STATE Initialize replay (experience) memory $\mathcal{H}=\emptyset$
\STATE Initialize action-value network $Q_{\theta}$ with random weights $\theta$
\STATE Initialize target action-value network $Q'_{\theta'}$ with weights $\theta'=\theta$
\STATE observe $s_0$
\FOR {$t=0$ to $T$}
    \STATE choose action $a_t = \mathop{\arg\max}\limits_{a} Q(s_t, a; \theta)$
    \STATE execute action $a_t$ and observe reward $r_t$ and state $s_{t+1}$
    \STATE store transition $(s_t, a_t, r_t, s_{t+1})$ in $\mathcal{H}$
        
    \FOR{$j=0$ to $N$}
        \STATE sample a batch of $K$ transitions $(s_{tj}, a_{tj}, r_{tj}, s_{t+1,j})$ from $\mathcal{H}$
        \STATE compute mean TD-error $f_j = \frac{1}{K} \sum_{k=1}^{K} (r_{tjk} + \gamma Q'(s_{t+1, jk}, \mathop{\arg\max}_a Q(s_{t+1,k}, a ; \theta') - Q(s_{tjk}, a_{tjk}; \theta))^2$
    \ENDFOR
    
    \STATE compute the Jacobian matrix $G=\nabla f$, where $f=(f_1, \ldots, f_N)^T$ 
    \STATE build and solve the quadratic programming (\ref{qp_dual1})-(\ref{qp_dual2}) and get a solution $\lambda$
    \STATE update $\theta \leftarrow \theta - \alpha G^T \lambda$
    \STATE from time to time copy weights into target network $\theta' \leftarrow \theta$
\ENDFOR
\end{algorithmic}
\end{algorithm*}

\section{Experiments}
We test the proposed method on four gym environments, named CartPole-v1, MountainCar-v0, LunarLander-v2 and Acrobot-v1.
In our experimental setting, we test DDQN, $\text{M}^2$DDQN with group size $N=5$ and $\text{M}^2$DDQN with group size $N=10$ and those three algorithms using the same configuration within each experiment.
We use a multi-layer fully connected neural networks to model the Q-network.
Furthermore, we use the cvxopt package\footnote{https://cvxopt.org/} to solve the quadratic programming.
Table \ref{parameters} shows the details of parameters.
The source code of $\text{M}^2$DDQN is available at \textbf{https://github.com/Myyura/Minimax-DQN}.

\begin{table}[htbp]
    \centering
    \resizebox{\linewidth}{!}{
    \begin{tabular}{|c|c|c|c|c|}
        \hline
        & CartPole & LunarLander & MountainCar & Acrobot \\
        \hline
        HiddenLayer & (128,64,64) & (128,64,64) & (64,32,32) & (64,32,32)\\
        LearningRate & 5e-4 & 5e-4 & 5e-4 & 5e-4 \\
        MaxStep & 200000 & 1000000 & 1000000 & 60000\\
        ReplaySize & 10000 & 50000 & 50000 & 3000\\
        BatchSize & 128 & 128 & 128 & 128\\
        Gamma & 0.99 & 0.99 & 0.99 & 0.99\\
        \hline
    \end{tabular}
    }
    \caption{Experimental Parameters}
    \label{parameters}
\end{table}

To measure the efficiency of training process, we focus on two evaluation metric.
Our main metric is \textit{learning speed}, in terms of the training step before the environment is considered  solved.
The learning speed shows whether the proposed method improves data efficiency or not.
Also, we need to ensure that the performance does not decrease when we use the max-mean loss.
Therefore, the second metric is \textit{quality of the best policy}, in terms of the maximum evaluation score of $50$ random games during the training process.
The gym leaderboard\footnote{https://github.com/openai/gym/wiki/Leaderboard} defines "solving" as getting average evaluation score of a specific value over $100$ random games.
In this paper, we follow the definition of gym leaderboard, but due to the limit of calculation resources, we defines "solving" as getting average evaluation score of a specific value over $50$ random games.

The normalized max score and the normalized training step before a environment is solved are showed in Table \ref{result_table}, with a more detailed learning curves and cumulative mean score curves on Figure \ref{mean_score} $\sim$ Figure \ref{train_curve}.

\begin{table}[htbp]
	\centering
	\setlength{\tabcolsep}{1mm}
	\begin{tabular}{*{7}{c}}
		\toprule
		Environment & Method & MaxScore & StepToSolve \\
		\midrule
		\multirow{3}*{CartPole-v1} & DDQN & 100\% & 100\% \\
		& $\text{M}^2$DDQN ($N=5$) & 100\% & \textbf{75.73\%}\\
		& $\text{M}^2$DDQN ($N=10$) & 100\% & 89.07\%\\
		\midrule
		\multirow{3}*{LunarLander-v2} & DDQN & 100\% & 100\% \\
		& $\text{M}^2$DDQN ($N=5$)& 105.13\% & 67.11\%\\
		& $\text{M}^2$DDQN ($N=10$)& \textbf{105.16}\% & \textbf{62.34\%}\\
		\midrule
		\multirow{3}*{MountainCar-v0} & DDQN & 100\% & 100\% \\
		& $\text{M}^2$DDQN ($N=5$)& 98.92\% & 52.77\%\\
		& $\text{M}^2$DDQN ($N=10$)& 99.86\% & \textbf{33.48\%}\\
		\midrule
		\multirow{3}*{Acrobot-v1} & DDQN & 100\% & - \\
		& $\text{M}^2$DDQN ($N=5$)& \textbf{106.04\%} & -\\
		& $\text{M}^2$DDQN ($N=10$)& 104.86\% & -\\
		\midrule
	\end{tabular}
	\caption{Normalized max score and normalized training step before a environment is solved (maximum evaluation score and steps to solve of DDQN is 100\%) on CartPole-v1, LunarLander-v2, MountainCar-v0 and Acrobot-v1. Note that Acrobot-v1 is an unsolved environment, which means it does not have a specified reward threshold at which it's considered solved.}
	\label{result_table}
\end{table}


In CartPole-v1 task, we define "solving" as getting average evaluation score of $495.0$ over $50$ random games.
For the quality of the best policy, all of three algorithms get a full evaluation score policy.
For the learning speed, the training steps before the environment is solved of DDQN, $\text{M}^2$DDQN ($N=5$) and $\text{M}^2$DDQN ($N=10$) are $100\%$, $75.73\%$ and $89.07\%$, respectively.
Proposed method learns slightly faster than standard DDQN.

In LunarLander-v2 task, we define "solving" as getting average evaluation score of $200.0$ over $50$ random games.
For the quality of the best policy, the maximum score of DDQN, $\text{M}^2$DDQN ($N=5$) and $\text{M}^2$DDQN ($N=10$) are $100\%$, $105.13\%$ and $105.16\%$, respectively.
For the learning speed, the training steps before the environment is solved of DDQN, $\text{M}^2$DDQN ($N=5$) and $\text{M}^2$DDQN ($N=10$) are $100\%$, $67.11\%$ and $62.34\%$, respectively.
Proposed method outperforms both in the quality of policy and learning speed.

In MountainCar-v0 task, we define "solving" as getting average evaluation score of $-110.0$ over $50$ random games.
For the quality of the best policy, the maximum score of DDQN, $\text{M}^2$DDQN ($N=5$) and $\text{M}^2$DDQN ($N=10$) are $100\%$, $98.92\%$ and $99.86\%$, respectively.
The maximum score of the three algorithms are almost the same.
For the learning speed, the training steps before the environment is solved of DDQN, $\text{M}^2$DDQN ($N=5$) and $\text{M}^2$DDQN ($N=10$) are $100\%$, $52.77\%$ and $33.48\%$, respectively.
Proposed method leads to a great improvement in the learning of this sparse reward environment.

In Acrobot-v1 task, there is no definition of "solving" in the gym leaderboard.
Therefore, we only focus on the quality of the best policy, the maximum score of DDQN, $\text{M}^2$DDQN ($N=5$) and $\text{M}^2$DDQN ($N=10$) are $100\%$, $106.04\%$ and $104.86\%$, respectively.
Proposed method get a better policy than standard DDQN.

In summary, the max-mean loss leads to a substantial improvement in all of the four tasks, especially in learning speed.
Also, the choice of group size $N$ may affect the performance of proposed method, in our experiments, larger group size of higher performance.


\begin{strip}
    \centering
    \begin{minipage}[t]{0.235\textwidth}
    \centering
    \includegraphics[width=3.85cm]{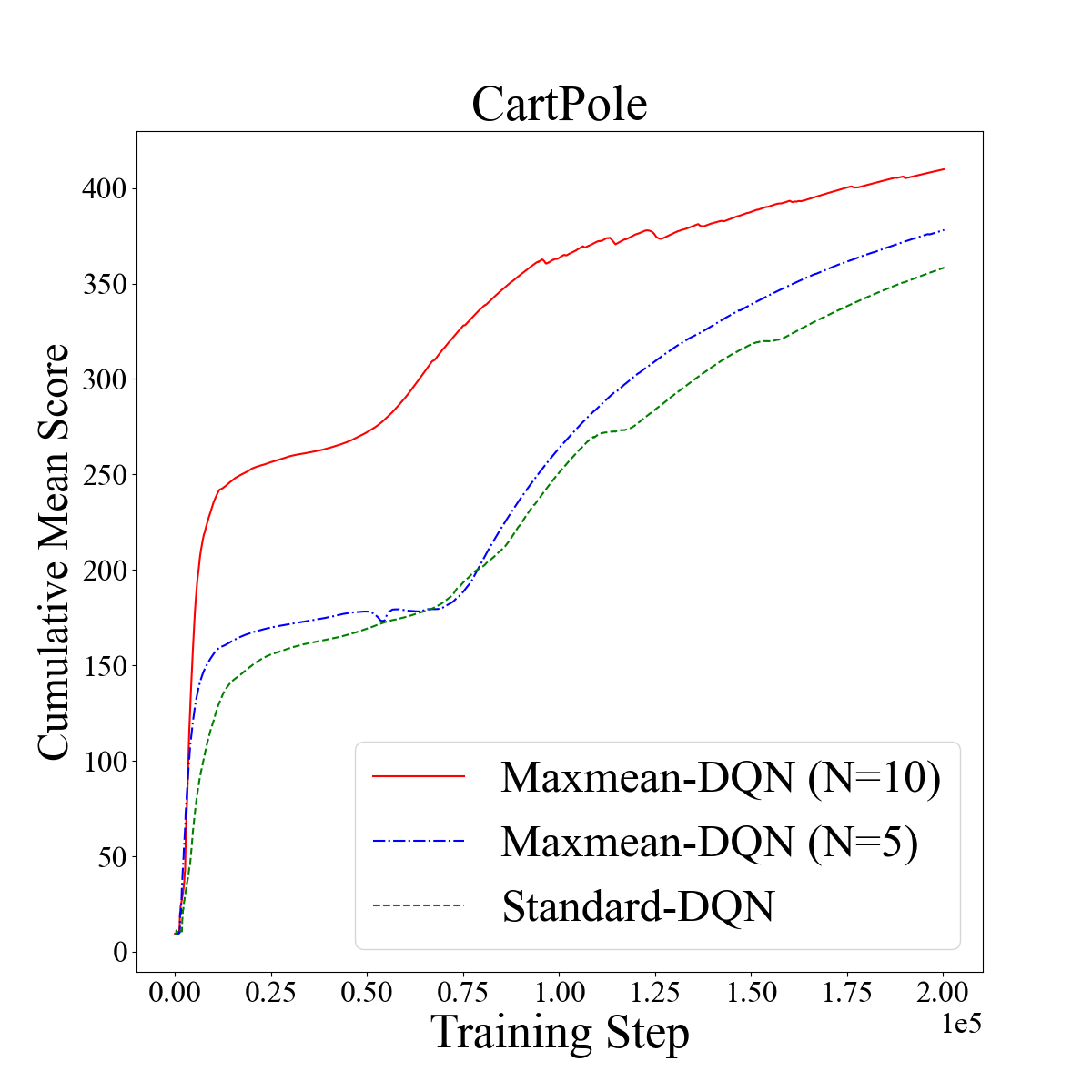}
    \label{cp_mean} 
    \end{minipage}
    \begin{minipage}[t]{0.235\textwidth}
    \centering
    \includegraphics[width=3.85cm]{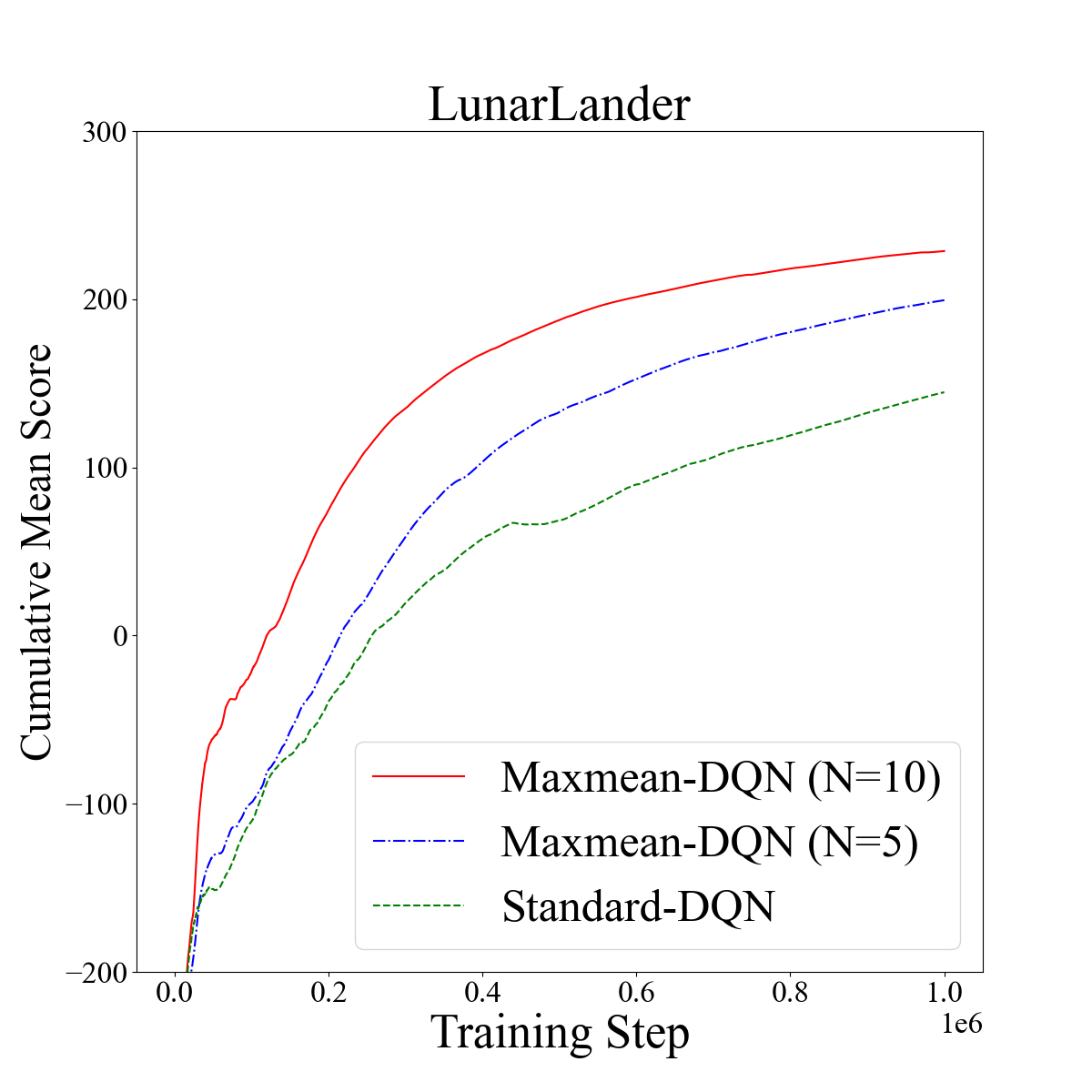}
    \label{ll_mean} 
    \end{minipage}
    \begin{minipage}[t]{0.235\textwidth}
    \centering
    \includegraphics[width=3.85cm]{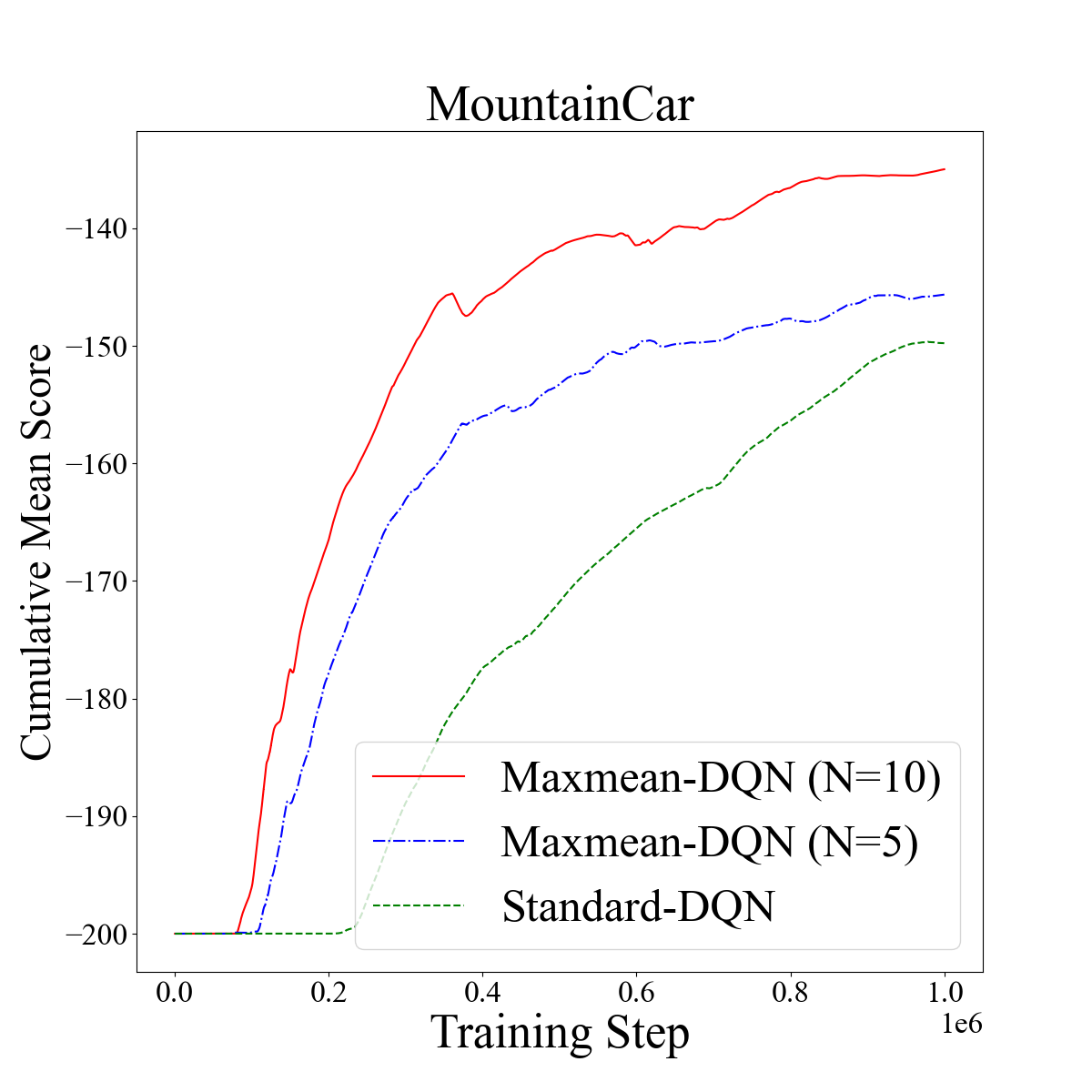}
    \label{mc_mean} 
    \end{minipage}
    \begin{minipage}[t]{0.235\textwidth}
    \includegraphics[width=3.85cm]{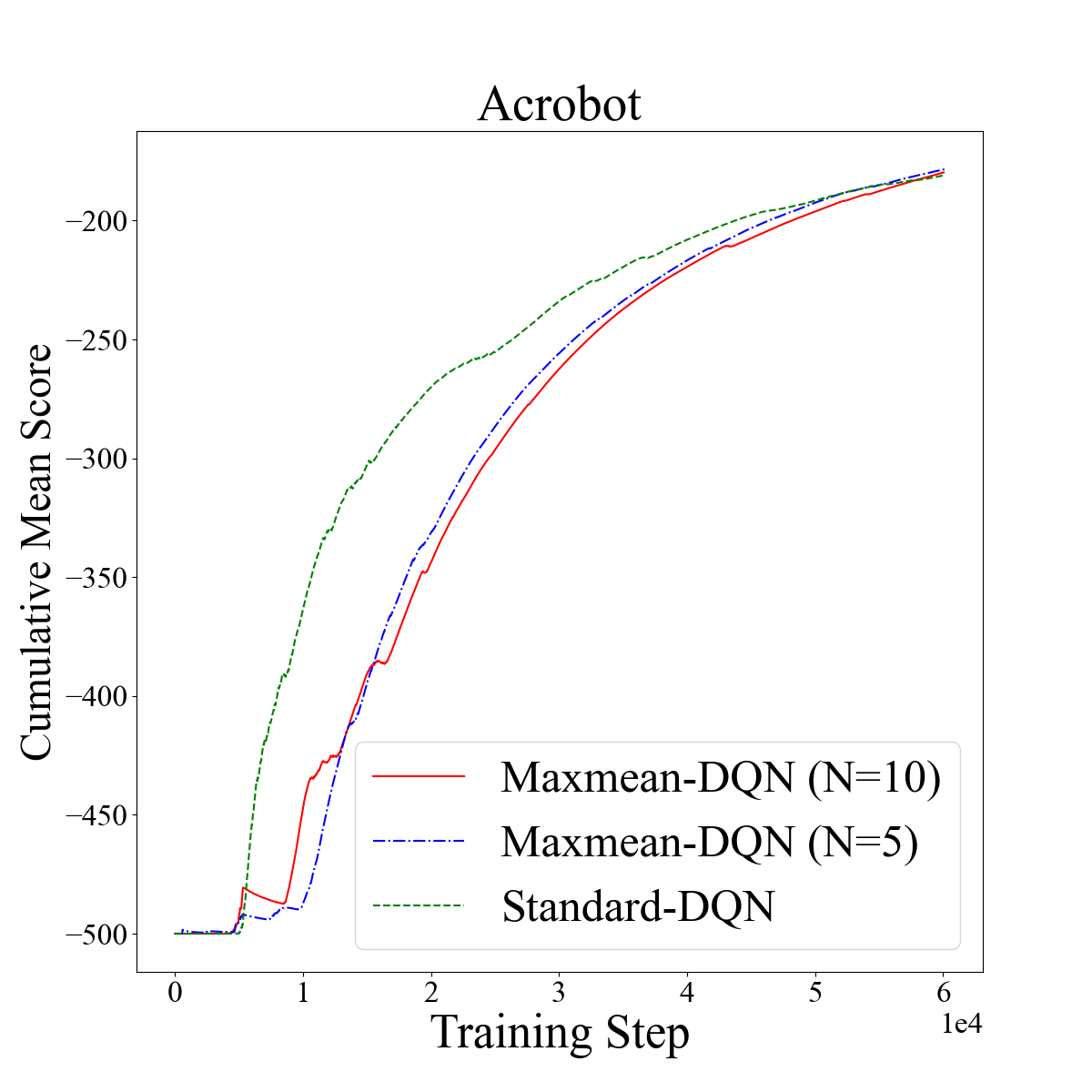}
    \label{ab_mean} 
    \end{minipage}
    \captionof{figure}{Cumulative mean evaluation score of DDQN and $\text{M}^2$DDQN with different group size $N$. Proposed method outperforms DDQN both in $N=5$ and $N=10$, especially in the MountainCar task, which is a sparse reward environment.}
    \label{mean_score}
\end{strip}

\section{Conclusion and Future Work}
In this paper, we propose a new framework of DQN to learn a policy more efficiently, by taking the max-mean loss instead of the standard loss.
This method can be combined with most of existing off-policy reinforcement learning algorithms by simply replacing the loss function.
We test proposed method with DDQN on four gym games (CartPole-v1, LunarLander-v2, MountainCar-v0 and Acrobot-v1).
The result show that max-mean loss speeds up learning by a factor $1.5 \sim 2$ and leads to a better policy in most of tested environments.

For future studies, we will first test the max-mean loss in other off-policy reinforcement learning algorithms, e.g., the learning of critic network in actor-critic based methods like Deep Deterministic Policy Gradient (DDPG) \cite{DDPG} and Soft Actor-Critic (SAC) \cite{SAC}.
Also, more detailed experiments on different group size $N$ should be performed, in order to find the relation between the performance and group size $N$. 

\begin{strip}
    \centering
    \begin{minipage}[t]{0.235\textwidth}
    \centering
    \includegraphics[width=3.85cm]{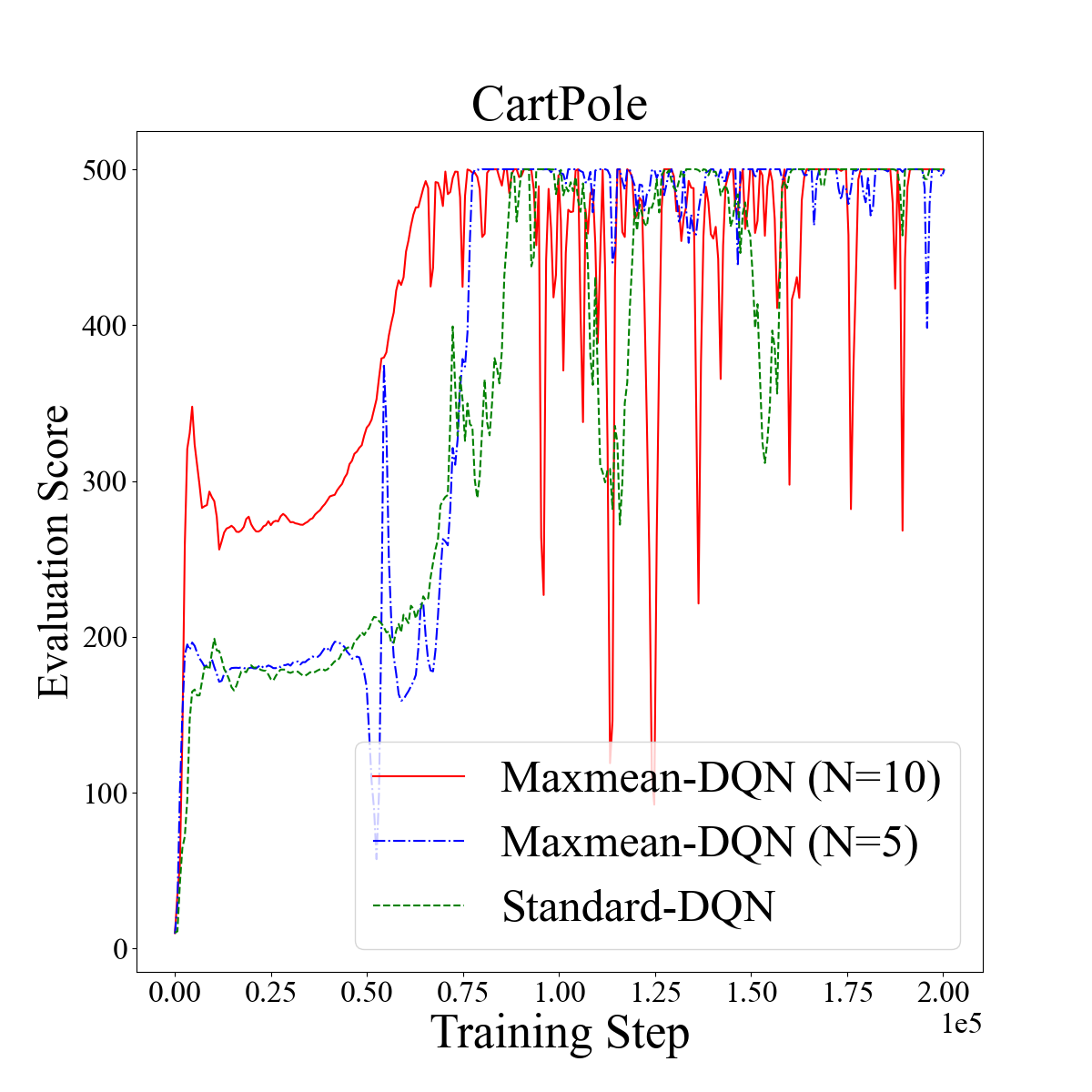}
    \label{cp_train} 
    \end{minipage}
    \begin{minipage}[t]{0.235\textwidth}
    \centering
    \includegraphics[width=3.85cm]{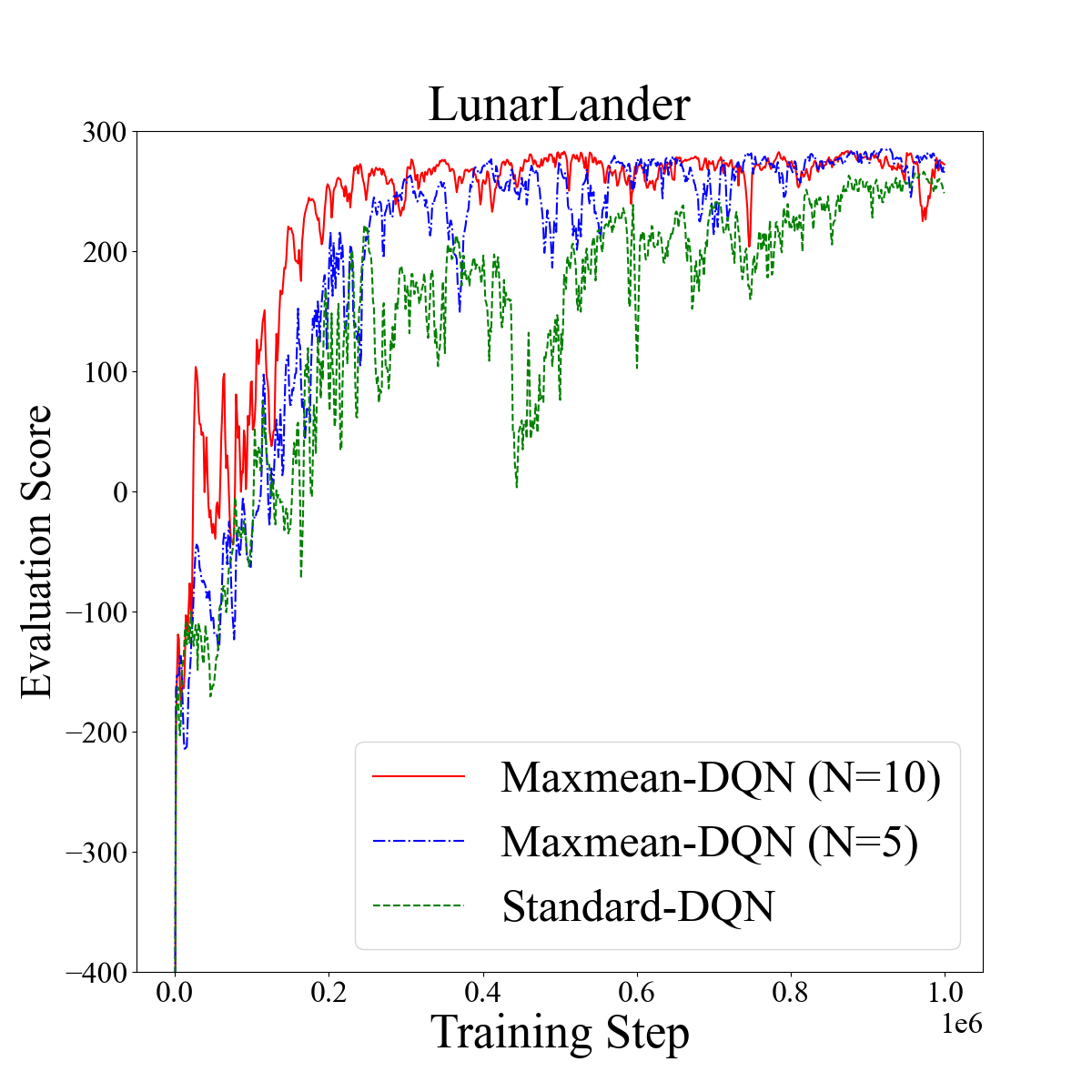}
    \label{ll_train} 
    \end{minipage}
    \begin{minipage}[t]{0.235\textwidth}
    \centering
    \includegraphics[width=3.85cm]{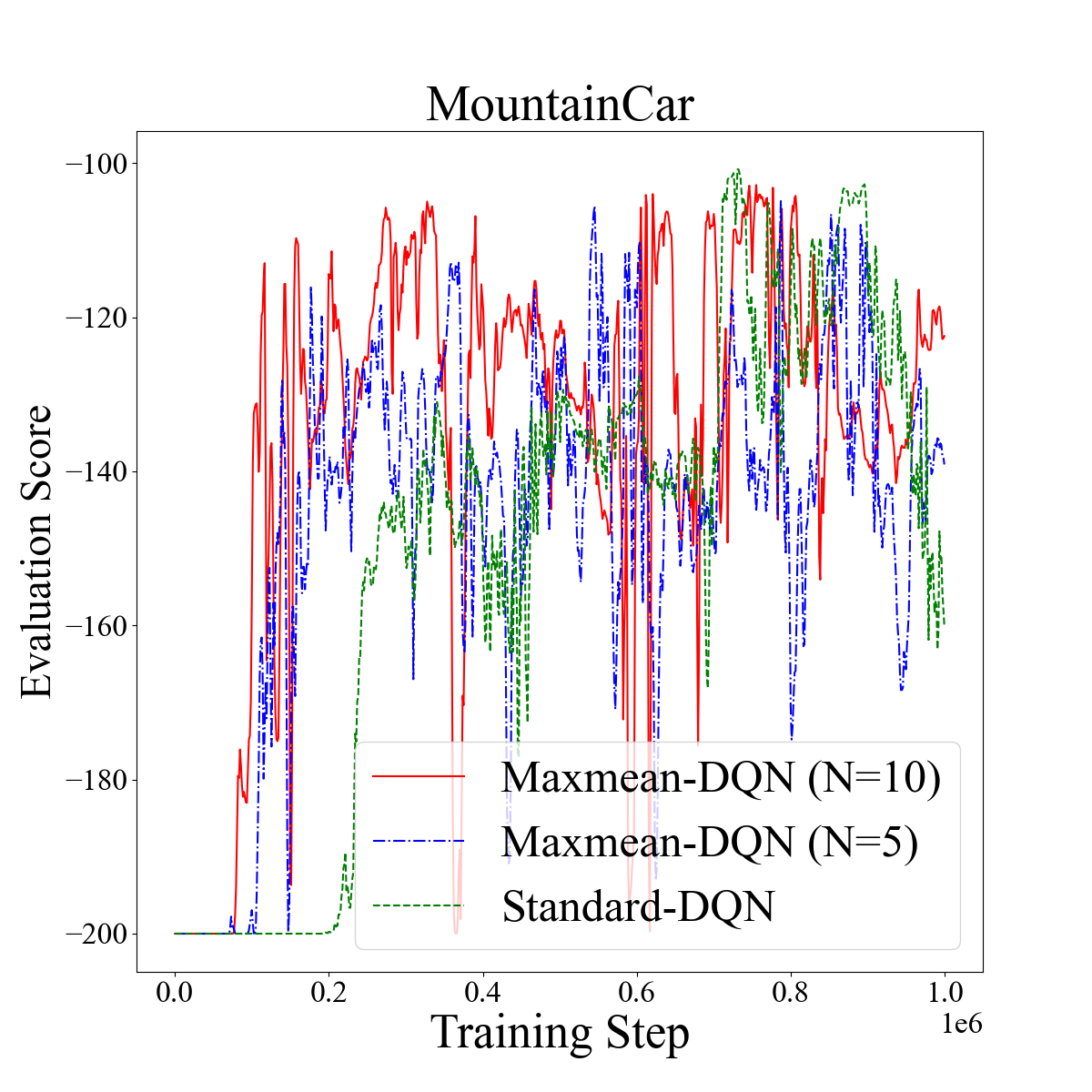}
    \label{mc_train} 
    \end{minipage}
    \begin{minipage}[t]{0.235\textwidth}
    \includegraphics[width=3.85cm]{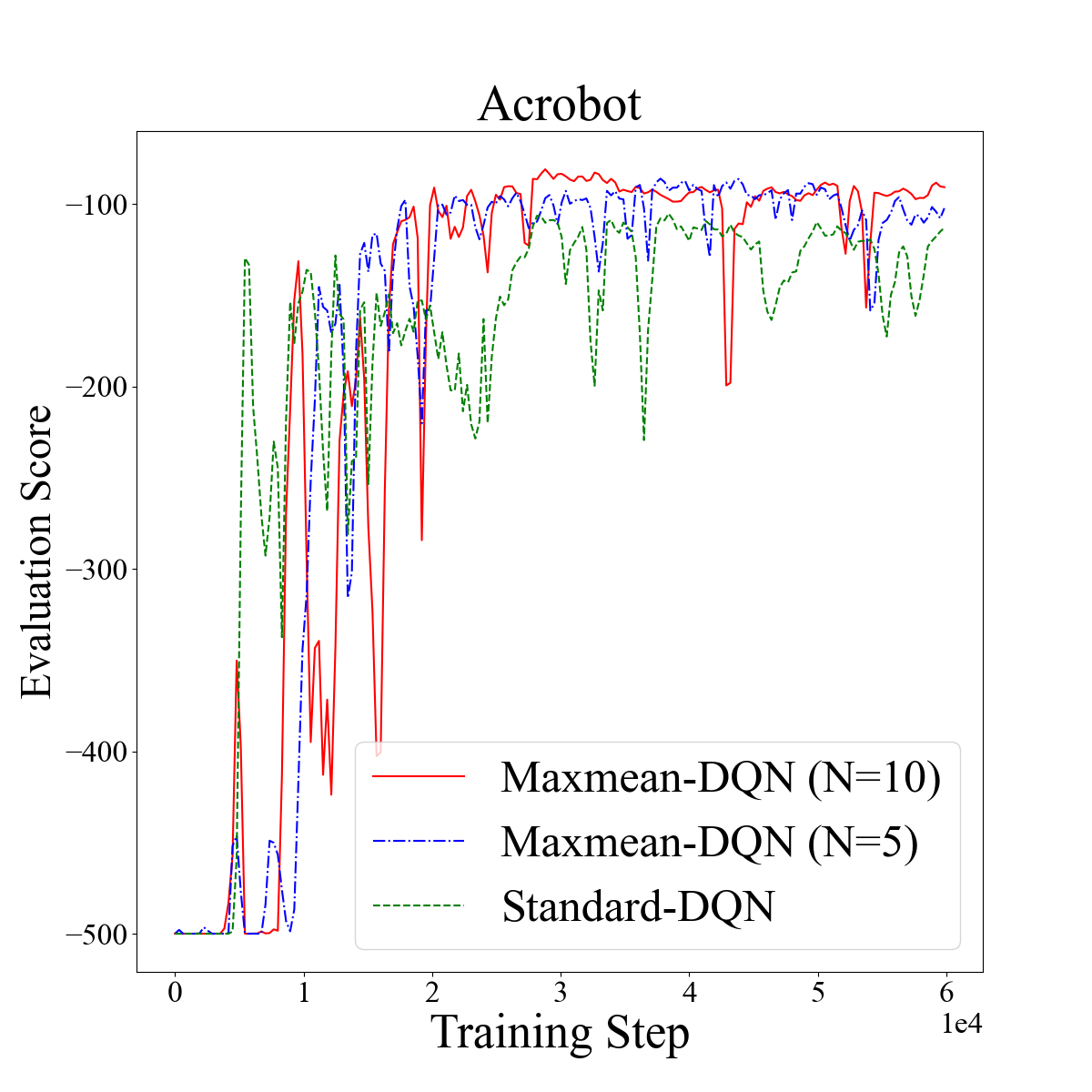}
    \label{ab_train} 
    \end{minipage}
    \captionof{figure}{Training curves of DDQN, $\text{M}^2$DDQN with different group size. The training curves show that proposed method also get a higher evaluation score in average.}
    \label{train_curve}
\end{strip}

\section*{Acknowledgment}

We thank Xuli Shen for useful discussions of the implementation of mini-max method.

\end{document}